\documentclass[11pt]{article}
\usepackage[utf8]{inputenc}
\usepackage{graphicx}
\usepackage{amsmath}
\usepackage{amssymb}
\usepackage{graphicx}
\usepackage{url}
\usepackage{multirow}
\usepackage{multicol}

\usepackage{lipsum}
\usepackage{setspace}
\usepackage{wrapfig}

\usepackage{geometry}

\usepackage[compact]{titlesec}
\titlespacing{\section}{0pt}{*0}{*0}
\titlespacing{\subsection}{0pt}{*0}{*0}
\titlespacing{\subsubsection}{0pt}{*0}{*0}
\titlespacing{\paragraph}{0pt}{*0}{*0}

\begin{document}


\doublespacing

\begin{center} \textbf{\Large Robust Explainability: A Tutorial on Gradient-Based Attribution Methods for Deep Neural Networks} \end{center}

\renewcommand{\thefootnote}{\fnsymbol{footnote}}

\begingroup\centering

Ian E. Nielsen$^{1}$%
\qquad Dimah Dera$^{1,2}$\\%
\qquad Ghulam Rasool$^{1}$%
\qquad Nidhal Bouaynaya$^{1}$%
\qquad Ravi P. Ramachandran$^{1}$%

$^{1}$ Rowan University%
\qquad $^{2}$ The University of Texas Rio Grande Valley%

\texttt{\{nielseni6, derad6, rasool, bouaynaya, ravi\}@rowan.edu}

\endgroup

\begin{abstract}

With the rise of deep neural networks, the challenge of explaining the predictions of these networks has become increasingly recognized. While many methods for explaining the decisions of deep neural networks exist, there is currently no consensus on how to evaluate them. On the other hand, robustness is a popular topic for deep learning research; however, it is hardly talked about in explainability until very recently. In this tutorial paper, we start by presenting gradient-based interpretability methods. These techniques use gradient signals to assign the burden of the decision on the input features. Later, we discuss how gradient-based methods can be evaluated for their robustness and the role that adversarial robustness plays in having meaningful explanations. We also discuss the limitations of gradient-based methods. Finally, we present the best practices and attributes that should be examined before choosing an explainability method. We conclude with the future directions for research in the area at the convergence of robustness and explainability.

\end{abstract}


\section{Introduction}

Deep learning (DL) has transformed the field of machine learning (ML) with deep neural networks (DNNs) being deployed in various real-world applications, including medical diagnosis, financial services, biometrics, intelligent transportation, social media, and smart home devices. Despite tremendous progress, their acceptance in mission-critical application areas is being hampered by two significant limitations. First, there is an inherent inability to explain decisions in a manner understandable to humans \cite{rudin2019stop}. Second, there is vulnerability to adversarial attacks, i.e., malicious and imperceptible alterations to the input, that can fool trained networks to alter their decisions drastically \cite{madry2018towards}. These seemingly two disparate concepts are intrinsically linked to each other and may have their origins in the data-driven nature of DNNs with a highly nonlinear input-output relationship and over-parameterized design. 

Explainability tackles the critical problem that human users cannot directly understand the complex behavior of DNNs or explain their underlying decision-making process. The explainability of ML models is the fundamental requirement for building trust with users and holds the key to their safe, fair, and successful deployment in real-world applications. The issue of explainability transcends the realm of scientific interest. The adoption of the General Data Protection Regulation (GDPR) by the European Union in May 2018 gives any citizen the ``right to explanation'' of an algorithmic decision made about them \cite{goodman2017european}. Explainability is both a legal right and a responsibility that has extensive social implications. The GDPR states that individuals ``have the right not to be subject to a decision based solely on automated processing''. 

Explainability in ML is not a new topic and has been handled in many different ways, including building interpretable models or generating post hoc explanations \cite{rudin2019stop}. This tutorial will focus on the latter. Given a trained neural network, either the input features are perturbed, and their effect on the network output is monitored, or a signal from the network output is back-propagated to the input. Either way, the resulting information from the perturbations or the gradient propagation provides an estimate of the contribution of input features to the output and can be presented as heatmaps.  

``How good is an explanation?'' is a fundamental question in explainability research. Generally, a visual analysis of the explanation is performed given the fact that these explanations are generated for humans to see and understand the behavior of the model. Recently, it has been shown that a visual analysis may not be a reliable method to ascertain the ``plausibility'' of an explanation \cite{adebayo2018sanity}. However, the lack of ground-truth explanation makes it challenging to quantitatively assess, compare, and contrast various explanations. 

A crucial property that all explainability methods should satisfy is insensitivity to minor input perturbations. That is, a small perturbation (possibly malicious) in the input, which does not affect network decision, should not significantly change the attributions \cite{alvarez2018robustness, ghorbani2019interpretation}. This notion of robustness is closely linked to reproducibility and replicability of explanations. Concurrently, the explanation of a decision should change significantly when the network is under an adversarial attack, that is, imperceptible malicious changes in the input that force the network to alter its decision \cite{madry2018towards}. In an ideal world, the attribution maps should be sensitive enough to detect adversarial attack and concurrently invariant to small perturbations in the input. 

This tutorial paper provides a thorough overview of gradient-based post hoc explainability methods, their attributional robustness, and the link between explainability maps and adversarial robustness. We restrict our focus on computer vision classification models, i.e., the networks are generally convolutional neural networks (CNNs) whose input data consist of images (e.g., from ImageNet datasets) and whose outputs are class scores or soft-max probabilities. We have created a website that provides links to all explainability methods discussed in the following sections, figures, and the code for generating these figures \footnote{\url{https://sites.google.com/view/robustexplainability/home}}. This tutorial is not meant to provide an exhaustive survey of all the explainability methods proposed for DL models.


\section{Taxonomy and Definitions}
Interpretability is defined as the ability to attach human-understandable meaning to the prediction of a model. Interpretability is a passive characteristic of a model that may refer to the level at which a given model makes sense to a human \cite{arrieta2020explainable}. On the other hand, explainability can be viewed as an active characteristic of an ML model. Explainability may consist of actions or procedures, usually performed by using a set of mathematical operations referred to as the explainability methods, to clarify or detail the internal functions of the ML model for a human observer \cite{arrieta2020explainable, doran2017does}. The explainability method must be an accurate proxy of the model's decision-making process and be comprehensible to humans. Reproducibility refers to obtaining consistent results using the same methods (experimental protocols, data, and code) as the original study \cite{national2019reproducibility}. Replicability is concerned with obtaining consistent results across studies aimed at answering the same scientific question using new data or methods \cite{national2019reproducibility}. Along with reproducibility and replicability, interpretability falls under the larger umbrella of explainability for ML models. However, the terms interpretability and explainability are interchangeably used in the literature. 

\begin{figure*}[tpb]
\centering
\centerline{\includegraphics[width=15cm]{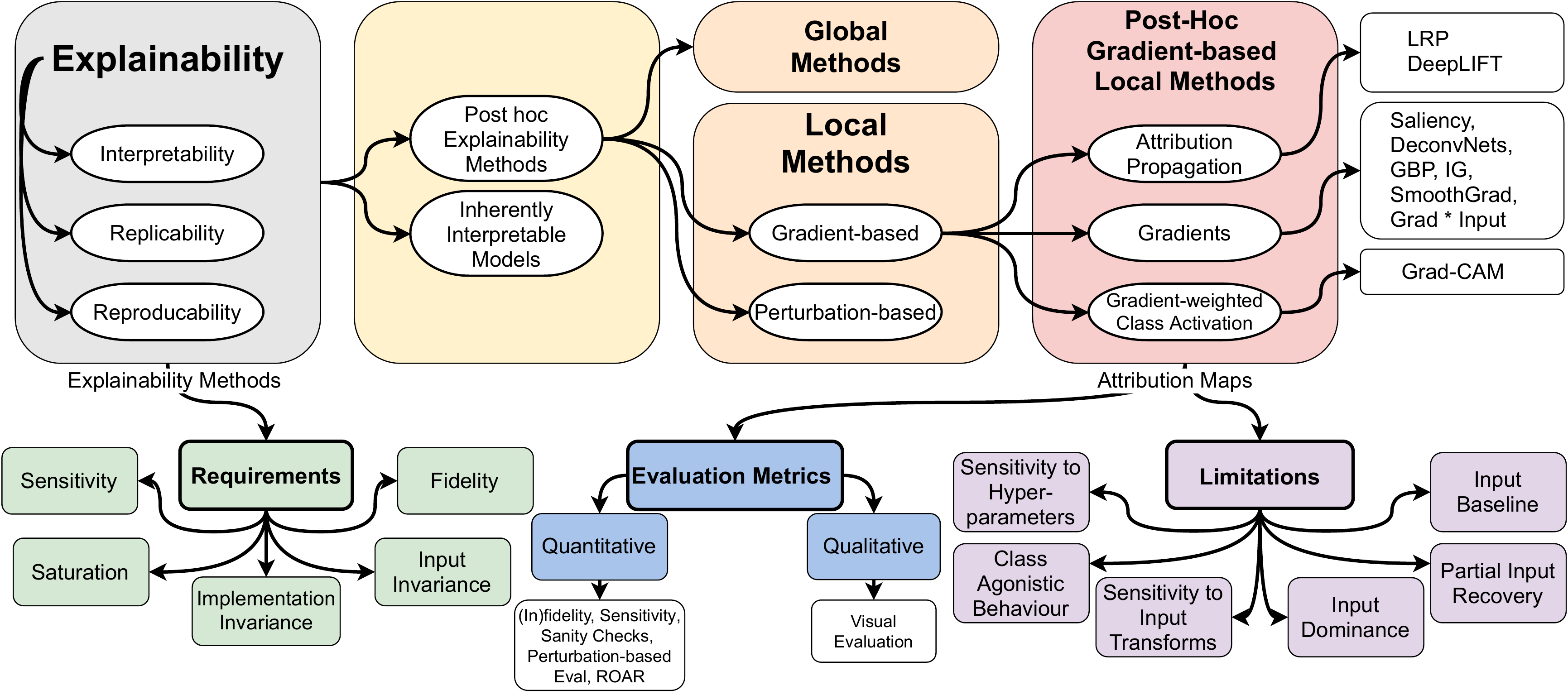}}
\caption{Visual breakdown of various explainability concepts discussed in the tutorial.}
\label{fig:full_diagram}
\end{figure*}

In Fig. \ref{fig:full_diagram}, we provide an overview of explainability concepts that are discussed in this tutorial paper. We outline various categories of explainability methods in ML (i.e., inherent vs. post hoc, global vs. local, and perturbation vs. gradient), requirements from these methods, various ways to evaluate these, i.e., metrics, and limitations of current methods used in the literature.  

\paragraph{Inherently Interpretable Models vs. Post hoc Explainability Methods}:~Some ML models are built to be inherently interpretable in the first place, e.g., linear models or decision trees \cite{rudin2019stop}. Other models, referred to as \emph{black box}, may require additional mathematical frameworks to explain their behavior to an audience targeting various use cases, e.g., understanding the model, debugging, providing explanations for legal purposes, or helping in decision making for downstream tasks. These mathematical frameworks, designed to explain black box models in post hoc settings, have their limitations and challenges over and above those in building ML models.  

\paragraph{Global vs. Local Explanations}:~Based on the scope and the purpose of explanation, a user can employ a local or a global interpretability method. Global interpretability methods attempt to explain the overall decision-making process of the model, i.e., how the inputs are transformed into the output decisions at the model level. These may be more useful to researchers and engineers trying to understand their models. In contrast, local interpretability methods attempt to explain specific decisions, i.e., what features of the input (e.g., pixels of an image) may have contributed (positively or negatively) to the model's output. 

The local interpretability problem can be formulated as estimating a number for each input feature that captures the effect of change in the feature value on the network output. The estimated numbers are presented as heatmaps and have the same dimension as the input features. In the literature, the terms, \emph{attribution}, \emph{relevance}, \emph{importance}, \emph{contribution}, \emph{sensitivity}, and \emph{saliency} scores are synonymously used.

\paragraph{Perturbation-based Methods vs. Gradient-based Methods}:~Various methods for post hoc local interpretability have been proposed. Two broad categories exist, namely, methods based on feature perturbation and others based on gradient information \cite{ancona2019gradient}. The former class of methods perturb input features (or a set of features) by masking or altering their values, and record the effect of these changes on the network performance. This requires multiple passes through the network to determine the importance of each input pixel, making perturbation-based attribution very computationally intensive. In the latter case, the gradients of the output (logits or soft-max probabilities) with respect to the extracted features or the input are calculated via backpropagation and are used to estimate attribution scores. Generally, the gradients are noisy, leading to attribution maps that may show contributions from irrelevant features. Various alterations to the gradient-based approach have been proposed to handle the challenge of noise in attribution maps \cite{smilkov2017smoothgrad}. Gradient-based approaches do not directly measure the affect of perturbing input features, and measuring the validity of methods is an ongoing challenge.

In Table \ref{Tab:Input-Outputs}, we provide a list of broader categories of explainability methods, the inputs that are required to use these methods, and the outputs that they provide.

\begin{table}[]
\begin{tabular}{lll}
\hline \hline
\textbf{\begin{tabular}[c]{@{}l@{}}Explainability Method\end{tabular}} & \multicolumn{1}{c}{\textbf{Input}} & \multicolumn{1}{c}{\textbf{Output}} \\ \hline \hline
\begin{tabular}[c]{@{}l@{}}Inherently interpretable\\ models\end{tabular} &
  \begin{tabular}[c]{@{}l@{}}Whole dataset\\ Single test sample\end{tabular} &
  \begin{tabular}[c]{@{}l@{}}Explanation of the model\\ and the prediction\end{tabular} \\ \hline
Global methods                                                            & Whole dataset                      & Explanation of the model            \\ \hline
\begin{tabular}[c]{@{}l@{}}Local methods - \\ Gradient-based\end{tabular} &
  \begin{tabular}[c]{@{}l@{}}Single test sample\\ Model information for \\ gradient back-propagation\end{tabular} &
  Explanation of the prediction \\ \hline
\begin{tabular}[c]{@{}l@{}}Local Methods - \\ Perturbation-based\end{tabular} &
  \begin{tabular}[c]{@{}l@{}}Single test sample\\ Perturbation pattern\end{tabular} &
  Explanation of the prediction \\ \hline \hline
\end{tabular}
\label{Tab:Input-Outputs}
\caption{Explainability methods, their inputs and outputs.}
\end{table}



\subsection{Requirements from Attribution Maps} \label{sec:Requirements}
Before we go into a detailed discussion about how to create local gradient-based post hoc attribution maps, it is relevant to consider what we expect from these attribution maps. Some of these requirements are defined axiomatically \cite{sundararajan2017axiomatic}.  

\paragraph{Implementation Invariance}:~As we know, ML models can be expressed and implemented in many different ways, mathematically or programmatically; however, two \emph{functionally equivalent} models should produce similar output for the same input. The attribution methods must be \emph{implementation invariant}, i.e., produce the same attribution scores for the same inputs on functionally equivalent networks, regardless of how these networks are implemented \cite{sundararajan2017axiomatic}. 

\paragraph{Input Invariance}:~Given the fact that neural networks are invariant to certain input transformations (e.g., a constant shift in the input), an attribution method must also be insensitive to such input transformations \cite{kindermans2019reliability}.



\paragraph{Fidelity}:~The fidelity or selectivity of an attribution method is linked with its ability to identify feature relevance. An attribution method with high fidelity assigns high attribution score to features that, when removed, greatly reduce network performance and vice versa \cite{tomsett2020sanity}.


\paragraph{Saturation}:~In the forward pass, an input feature may saturate the network, owing to the nonlinear activation function being used, e.g., the rectified linear unit (ReLU) function \cite{sundararajan2017axiomatic}. Consider a neural network with one ReLU, $f(x) = 1 - \text{ReLU}(1-x)$. For all input values $x>1$, we have: $f(x) = 1$ and $\frac{d}{dx} f(x) = 0$. Despite the fact that the input feature may change significantly, the function output stays the same and the gradient remains at zero. An attribution method must tackle the saturation in the network while estimating attributions. One possible method is to use a \emph{reference} input or \emph{baseline}, which can be zero (black pixel), a random number, or an average value calculated over the input dataset.

\paragraph{Sensitivity}:~Considering a neutral baseline (e.g., zero or black image) for all features, \emph{sensitivity} requires that the output of the model for an input should be decomposable as the sum of the individual contributions from the input features \cite{ancona2019gradient}. This property is also referred to as \emph{completeness} or \emph{summation to delta} \cite{sundararajan2017axiomatic, shrikumar2017learning}. For an attribution method to be considered sensitive, it must assign a non-zero attribution score to the single distinctive feature between two similar inputs. Furthermore, sensitivity requires that any feature, which does not affect the output of the network, must be given a zero attribution score \cite{sundararajan2017axiomatic}.





\section{Gradient-Based Attribution Methods}
This section introduces various methods that use gradient information in post hoc settings and build attribution maps. These methods use the gradient of the network output (logits or soft-max probabilities) with respect to the input features.


We consider a network with an $N$-dimensional input $\mathbf{x} = \{x_i\}_{i=1}^N \in \mathbb{R}^N$ and a $C$-dimensional output $\mathbf{S(x)} = \{S_c\}_{c=1}^C \in \mathbb{R}^C $, where $C$ is the total number of classes and $S_c (\mathbf{x})$ represents the network's score function. Note that $S_c$ can be either a class score (logit) or soft-max probability. We use the term ``gradient'' for $\frac{\partial}{\partial \mathbf{x}} S_c (\mathbf{x}) $. The goal of attribution methods is to estimate the attribution map, $\mathbf{A}^c = \{A_i^c\}_{i=1}^N \in \mathbb{R}^N$. The attribution map $\mathbf{A}^c$ captures the importance of each input feature for a specific output class $c$. In computer vision applications, we consider CNNs with image inputs, i.e., the pixels of the image are considered input features. The resulting attribution map $\mathbf{A}^c $ has the same size $N$ as the input. 

\subsection{Gradients}
We consider a linear model with $N+1$ parameters $\boldsymbol{\theta}$ and an $N$ feature input,
\setlength\abovedisplayskip{0.5pt}
\setlength\belowdisplayskip{2.0pt}
\begin{align} \label{eqn:linear_model}
    y &= \theta_0 + \theta_1 x_1 + \ldots + \theta_N x_N + \epsilon = \boldsymbol{\theta}^T \mathbf{x} + \epsilon,
\end{align}
where $\epsilon$ is the modeling error and $\theta_0$ is the bias. 
The partial derivative of the output $y$ with respect to the input $\mathbf{x}$ results in model parameters $\boldsymbol{\theta}$, which represent contributions of input features. Thus, for the linear case, model parameters serve as feature attributions. Recently, gradients have also been used as features for perceptual image quality assessment and out-of-distribution classification \cite{kwon2019distorted}.

\paragraph{Saliency Maps}:~Simonyan \emph{et al}. used a similar formulation with an absolute value for constructing \emph{Saliency maps} for DNNs, $\mathbf{A}^c_{\text{Saliency}} = \left| \frac{\partial S_c}{\partial \mathbf{x}} \right|_{\mathbf{x}_0}$, where ${\mathbf{x}_0}$ is the input \cite{simonyan2013deep}. It is important to highlight that $S_c$ is a nonlinear function of the input $\mathbf{x}$ and thus, in contrast to the linear case, the model parameters no more represent feature attributions. It has been shown that saliency maps represent the first-order approximation of the attributions \cite{simonyan2013deep}. The major challenge with saliency maps is that they are visually noisy and a great deal of research has focused on removing noise and improving visualization \cite{smilkov2017smoothgrad}. 

\paragraph{Deconvolutional Networks (DeconvNets)}:~Saliency maps are closely related to DeconvNets proposed by Zeiler and Fergus \cite{zeiler2014visualizing}. In saliency maps, the gradient signals are zeroed during backpropagation at each ReLU when the input to the same ReLU was negative during forward pass. In contrast, DeconvNet reduces the negative gradients to zero at each ReLU, ignoring the fact whether the input to the same ReLU was negative or positive during the forward propagation. 

\paragraph{Guided Backpropagation (GBP)}:~GBP combines operations from both saliency maps and DeconvNet \cite{springenberg2014striving-GBP}. That is, during backpropagation, the attribution signal is reduced to zero at a ReLU when either the gradient signal itself is negative or the input to the ReLU at the time of the forward pass was negative. Removing negatively contributing features may reduce noise and improve visualization of attribution maps in some cases.

\paragraph{SmoothGrad (SG)}:~SmoothGrad reduces noise and visual diffusion by averaging over explanations generated for multiple noisy copies of the input $\mathbf{x}$ \cite{smilkov2017smoothgrad}. For a saliency map $\mathbf{A}^c$ calculated for the input $\mathbf{x}$, SmoothGrad is given by $\mathbf{A}_{\text{sg}}^c = \frac{1}{n} \sum_{i=1}^{n} \mathbf{A}^c \left(\mathbf{x} + \mathcal{N} (0, \sigma^2)\right)$, where $n$ is the number of samples and $\mathcal{N}$ represents the Gaussian distribution. 

\paragraph{Gradient$\odot$Input}:~In Gradient$\odot$Input, the attribution scores are calculated by element-wise multiplication of gradients with the input, i.e., $\mathbf{A}^c_{\text{Gradient}\odot \text{Input}} = \frac{\partial S_c(\mathbf{x})}{\partial \mathbf{x}} \odot \mathbf{x}$.
The element-wise multiplication can be considered as an application of a model-independent filter (the input), which may reduce noise and smoothen the attribution maps \cite{ancona2019gradient}.



\paragraph{Integrated Gradients (IG)}:~IG can be considered a smoother version of Gradient$\odot$Input, specifically designed to satisfy two axioms of explainability, i.e., sensitivity and implementation invariance \cite{sundararajan2017axiomatic}. IG along the $i^{th}$ dimension for an input $x$ and baseline $\acute{x}$ is given by $\text{IG}_i (x) = (x - \acute{x}) \int_{0}^1 \frac{\partial}{\partial{x}_i} S_c \left(\acute{x} + \alpha (x - \acute{x})\right) d\alpha$.
IG calculates the average of all gradients along a straight line between the baseline and the input. In practice, we can only use a finite number of samples to approximate the integral, which may introduce an approximation error.



\subsection{Attribution Propagation}
Attribution propagation can be considered as an alternative to calculating gradients. Recursively, attribution propagation methods decompose the decision made by the network into contributions from previous layers, all the way to the input. These methods use forward-pass activations (starting with the activation of the neuron in the last layer) to move back layer-by-layer in the network and distribute the burden of the decision over the input features. This class of methods includes various forms of Layer-wise Relevance Propagation (LRP), Deep Taylor Decomposition \cite{montavon2017explaining}, and Deep Learning Important FeaTures (DeepLIFT) \cite{shrikumar2017learning}. These methods do not strictly use \emph{gradients} intrinsically; however, their relationship to Gradient$\odot$Input has been mathematically established \cite{ancona2019gradient}.    

\paragraph{Layer-Wise Relevance Propagation (LRP)}:~LRP propagates relevance scores from the last layer of the network to the input using the ``conservation property'' \cite{bach2015LRP}. That is, what was received by a neuron in the forward pass (activations) must be redistributed to the lower layer (a layer nearer to the input) by an equal amount. Going from the output to the input, layer-by-layer, the relevance scores are scaled at each layer using the information from the forward pass. LRP starts with the activation of the neuron in the last layer. Let $j$ and $k$ be neurons at two consecutive layers $l$ and $l+1$, with layer $l$ closer to the input. Let $\theta_{jk}$ be the learnable parameters that connect both layers. The neuronal activation $a_{k}^{[l+1]}$ in the forward pass is defined as $a_k^{[l+1]} = \text{ReLU} \left( \sum_j a_j^{[l]} \theta_{jk}\right) $. Given that we have $r_k^{[l+1]}$, i.e., relevance score at layer $l+1$, we can calculate the relevance score $r_{j}^{[l]}$ at layer $l$ using:
\setlength\abovedisplayskip{3.0pt}
\setlength\belowdisplayskip{6.0pt}
\begin{align} \label{eqn:LRP}
    r_j^{[l]} = \sum_k \frac{a_j^{[l]} \theta_{jk}}{\epsilon + \sum_j a_j^{[l]} \theta_{jk}} r_k^{[l+1]}.
\end{align}
Equation \ref{eqn:LRP} is called LRP-$\epsilon$, where $\epsilon$ is added to absorb some relevance when the contributions to the activation of neuron $k$ are weak or contradictory. When $\epsilon \gg \sum_j a_j^{[l]} \theta_{jk}$, only the most salient explanation factors survive the absorption, leading to noise reduction and sparser explanations \cite{bach2015LRP}. It has been shown that for CNNs with ReLU activation functions, LRP-$\epsilon$ implements a slightly modified form of Gradient$\odot$Input, where the gradients are normalized at each layer by the activations $\sum_j a_j^{[l]} \theta_{jk}$ \cite{ancona2019gradient}. Montavon \emph{et al}. proposed Deep Taylor Decomposition, which provided theoretical foundations for LRP using Taylor series approximation \cite{montavon2017explaining}.




\paragraph{Deep Learning Important FeaTures (DeepLIFT)}:~DeepLIFT was designed to tackle the saturation problem using ``reference activations'', calculated in the forward pass with the baseline input \cite{shrikumar2017learning}. DeepLIFT compares the activation of each neuron to its reference activation and assigns contribution scores according to the difference \cite{shrikumar2017learning}. It has been shown that DeepLIFT (Rescale rule) is equivalent to Gradient$\odot$Input \cite{ancona2019gradient}. 

\subsection{Gradient-weighted Class Activation Mapping (Grad-CAM)}
\begin{figure*}[h]
\centering
\centerline{\includegraphics[width=14cm]{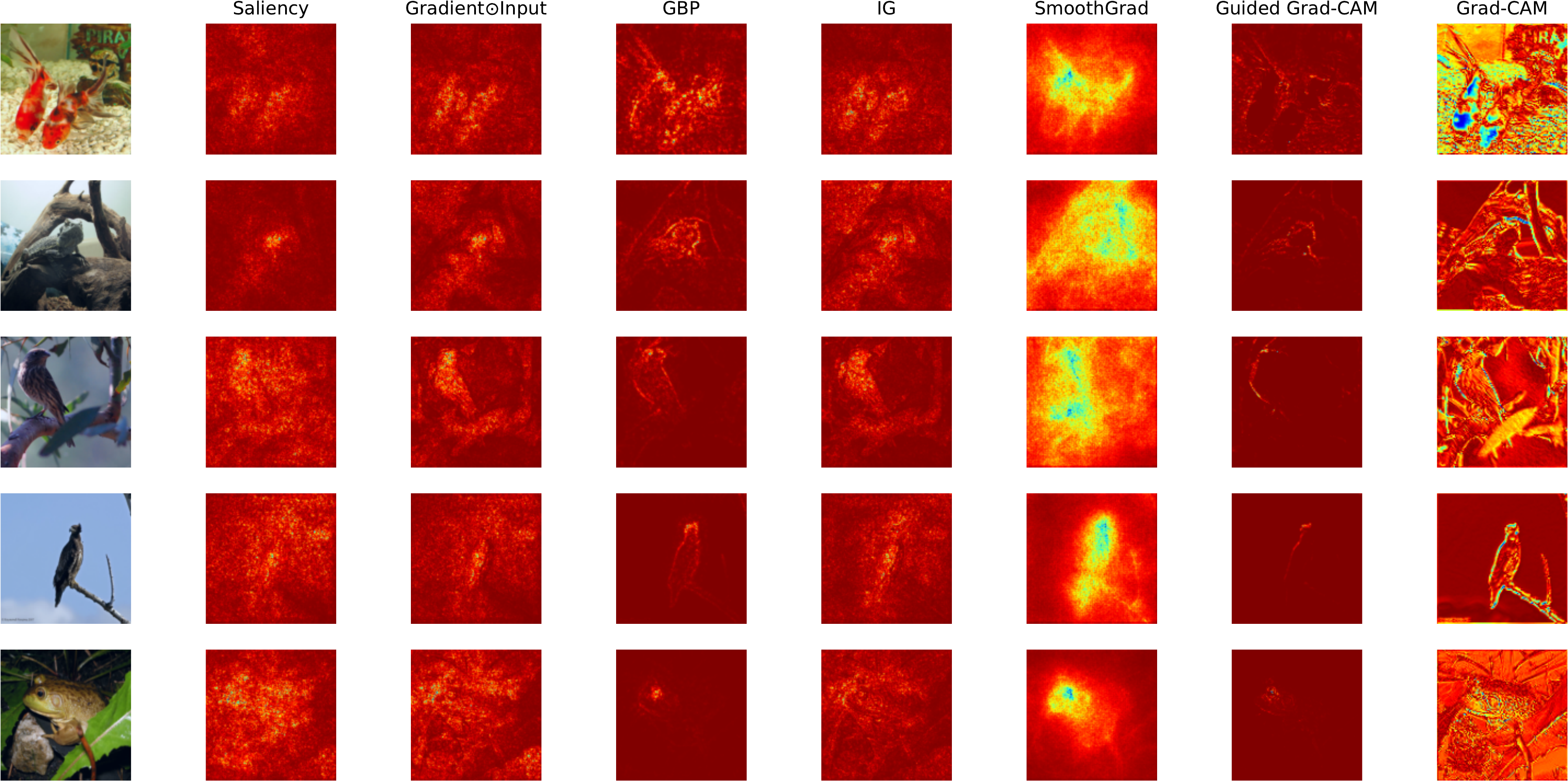}}
\caption{Attribution maps generated using different methods are presented. The first column presents test images from ImageNet and rest of the columns present attribution maps estimated using various methods. Abbreviations used: GBP - Guided Backpropagation, IG - Integrated Gradients, and Grad-CAM - Gradient-weighted Class Activation Mapping.}
\label{fig:attribution_maps}
\end{figure*}

Grad-CAM uses the class-specific gradient information flowing into the final convolutional layer of a CNN to produce a coarse localization map of the important features of the input \cite{selvaraju2017grad}. Grad-CAM analyzes which regions are activated in the feature maps of the last convolutional layer. Grad-CAM can be combined with GBP, referred to as Guided Grad-CAM, to improve pixel-level granularity of attribution maps. In Fig. \ref{fig:attribution_maps}, we present attribution maps generated using various methods.

Recently, Grad-CAM have been extended for building contrastive explanations, thus answering the question of why a prediction $P$ was produced by the network rather than a different prediction $Q$ \cite{prabhushankar2020contrastive, prabhushankar2021contrastive}.  

Another approach uses explainability maps along with logical prior knowledge during the training process to produce more interpretable explanations \cite{diaz2022explainable}.


\section{Analysis of Gradient-Based Attribution Methods}

Attribution maps are designed to explain the decisions made by ML models. However, a large body of research has found that various approaches to create these attributions have their own limitations \cite{kindermans2019reliability, adebayo2018sanity, ghorbani2019interpretation}. In Table \ref{Tab:requirements}, we provide a list of various gradient-based explainability methods and the requirements (as provided in Section \ref{sec:Requirements}) that they satisfy. 
%
Starting with measures that can be used to evaluate explanations in DL models, we provide a detailed analysis of the performance of these methods and their limitations. 


\subsection{Evaluation of Attribution Maps}
``How good is an explanation?'' is one of the fundamental questions in ML explainability research. The lack of ground truth explanations makes validating attribution methods challenging. Ideal evaluation of these methods will depend upon fully knowing the process of how the ML model reached its decision - the very problem that we are trying to solve. Furthermore, it is hard to disentangle the errors made by models from the errors made by attribution methods \cite{sundararajan2017axiomatic}. Given that the notion of explanation is centered around human visual perception, the predominant evaluations of attributions have been subjective. However, objective evaluation is equally, or perhaps more, important to establish rigorous theoretical foundations, compare and contrast various approaches, and improve upon these methods \cite{yeh2019fidelity}.



\begin{table}[htpb]
\begin{tabular}{lccccc}
\hline \hline
Requirements &
  \multirow{2}{*}{\begin{tabular}[c]{@{}l@{}}Implementation\\ Invariance\end{tabular}} &
  \multirow{2}{*}{\begin{tabular}[c]{@{}l@{}}Input\\ Invariance\end{tabular}} &
  \multirow{2}{*}{Fidelity} &
  \multirow{2}{*}{Saturation} &
  \multirow{2}{*}{Sensitivity} \\ \cline{1-1}
Methods &
   &
   &
   &
   &
   \\   \hline \hline
Saliency              & Yes            & Yes            & No                       & No         & Yes           \\ \hline
GBP                   & No             & Yes/No*        & Yes                      & Yes        & No            \\ \hline
SG                    & Yes            & Yes            & Yes                      & No         & Yes           \\ \hline
Grad$\odot$Input            & Yes            & No             & Yes                      & No         & Yes \\ \hline
IG                    & Yes            & No             & Yes                      & No         & Yes           \\ \hline
DeepLIFT              & No             & No             & ?$^\&$               & Yes        & ?$^\&$    \\ \hline
Grad-CAM              & Yes/No$^\$$       & Yes            & Yes                      & Yes/No$^\$$   & Yes           \\ \hline \hline
\end{tabular}
\label{Tab:requirements}
\caption{Various explainability methods and the requirements they satisfy. *depends on the network architecture. $^\$$depends on the layer chosen for attribution, $^\&$not tested. }
\end{table}

\paragraph{Visual Evaluation}:~A visual analysis of the attribution maps may seem to be the most plausible way of evaluation as these are created to explain the behavior of DL models to human operators and designers. Visual analysis includes qualitative displays of explanation examples, crowd-sourced evaluations of human satisfaction with the explanations, as well as whether humans are able to understand the model output \cite{yeh2019fidelity}. However, it may be misleading to rely solely on visual analysis for determining whether an attribution method is able to capture the features that a network considers important \cite{adebayo2018sanity}. A visual analysis may bias the evaluation of how humans understand the phenomenon and make decisions, rather than capturing how the network reached a particular decision. This may hold true especially for the methods that multiply the input with the gradient, i.e., Gradient$\odot$Input and IG. 

\paragraph{Feature Perturbation-based Evaluation}:~Removing the most important features identified by an attribution method and recording its effect on the performance of the network may provide an objective approach to evaluate various attribution methods \cite{samek2016evaluating}. A good attribution method will identify the most important pixels, which when removed should maximally degrade the network performance. The metric is referred to as the Most Relevant First (MoRF). As the network input size is fixed, the removed pixels are replaced with either the average value (calculated over the input dataset), zero (i.e., black pixel), or random values \cite{tomsett2020sanity}. It is obvious that replacing pixels with an average value or black pixels can introduce high-frequency edges, which may degrade network performance - unrelated to the removal of important pixels \cite{srinivas2019full}. We may choose to remove the least important pixels first - thus partially decoupling the effects of artifacts introduced by high-frequency edges from those caused by removing important pixels. The metric is referred to as the Least Relevant First (LeRF) \cite{srinivas2019full}. 

A recent study by Tomsett \emph{et al}. evaluated the reliability of both MoRF and LeRF using four different statistical tests from the psychometric literature \cite{tomsett2020sanity}. These tests included inter-rater reliability, inter-method reliability, internal consistency reliability, and test-retest reliability, where each image corresponded to a different rater and methods included different attribution map generation techniques. Both MoRF and LeRF showed: (1) high variance across all tested images, (2) sensitivity to whether the removed pixels were replaced with the mean of the dataset or random values, (3) low inter-rater reliability, i.e., the rankings of different methods were highly inconsistent, and (4) low correlation with each other. The results were reported for the classification task using the CIFAR-10 dataset. The absence of ground truth explainability and the limited testing (using one dataset only) makes it hard to generalize these results to other metrics. However, the study raised important questions about the validity of different metrics that are extensively used in the explainability literature.

\paragraph{Remove and Retrain (ROAR)}:~ In ROAR, the model is retrained and evaluated every time after removing a set of most important pixels \cite{hooker2019benchmark}. ROAR is computationally expensive and does not address the question of validity of explanation for each input, rather evaluates the method globally over the whole dataset. Furthermore, the retraining strategy may force the network to learn from the features that were not present in the original dataset (e.g., high-frequency edges introduced due to pixel replacement).

\paragraph{(In)fidelity and Sensitivity}:~(In)fidelity quantifies the statistically expected difference between (1) the dot product of the input perturbation to the attribution scores and (2) the output perturbation (difference in the score function $S_c(\mathbf{x})$ values after \emph{significant perturbations} introduced in the input $\mathbf{x}$) \cite{yeh2019fidelity}. (In)fidelity allows for a number of significant perturbations, including random and non-random perturbations that lead the input towards a predefined single or multiple baseline values. Random perturbations with a small amount of additive Gaussian noise allows the measure to be robust to small mis-specifications or noise in either the test input or the reference point \cite{yeh2019fidelity}. On the other hand, the ``sensitivity'' measures the degree to which the explanation is affected by \emph{insignificant perturbations} in the test point \cite{yeh2019fidelity}. A good attribution method will exhibit low sensitivity, i.e., producing same explanations for minor variations in the input.


\paragraph{Sanity Checks}:~Recently, Adebayo \emph{et al.} introduced two \emph{sanity checks} for evaluating the sensitivity of attribution methods to the model parameters and the dataset \cite{adebayo2018sanity}. The first check consists of replacing all the learned parameters of the network with random numbers. The resulting attribution maps are compared to the original maps using various correlation metrics to ascertain whether the attribution maps were able to capture the changes in the network parameters. The second check evaluates attribution methods on networks trained using randomly permuted labels. The attribution maps which remain unchanged for either of these checks are considered to have failed. 



\subsection{Limitations of Attribution Maps}
The attribution methods explain the behaviour of a model for a single test point selected from the evaluation dataset. The explanation provided by the attribution methods for the selected single point may be too brittle and could lead to a false conclusion about the performance of the model \cite{alvarez2018robustness}. Thus, understanding a complex model with a single or even multiple pointwise explanations without theoretically grounded metrics is perhaps too optimistic \cite{alvarez2018robustness}. Explaining a model at a single point and then generalizing to the whole dataset is an open question for the research community. 

\begin{figure*}[h]
\centering
\centerline{\includegraphics[width=12cm]{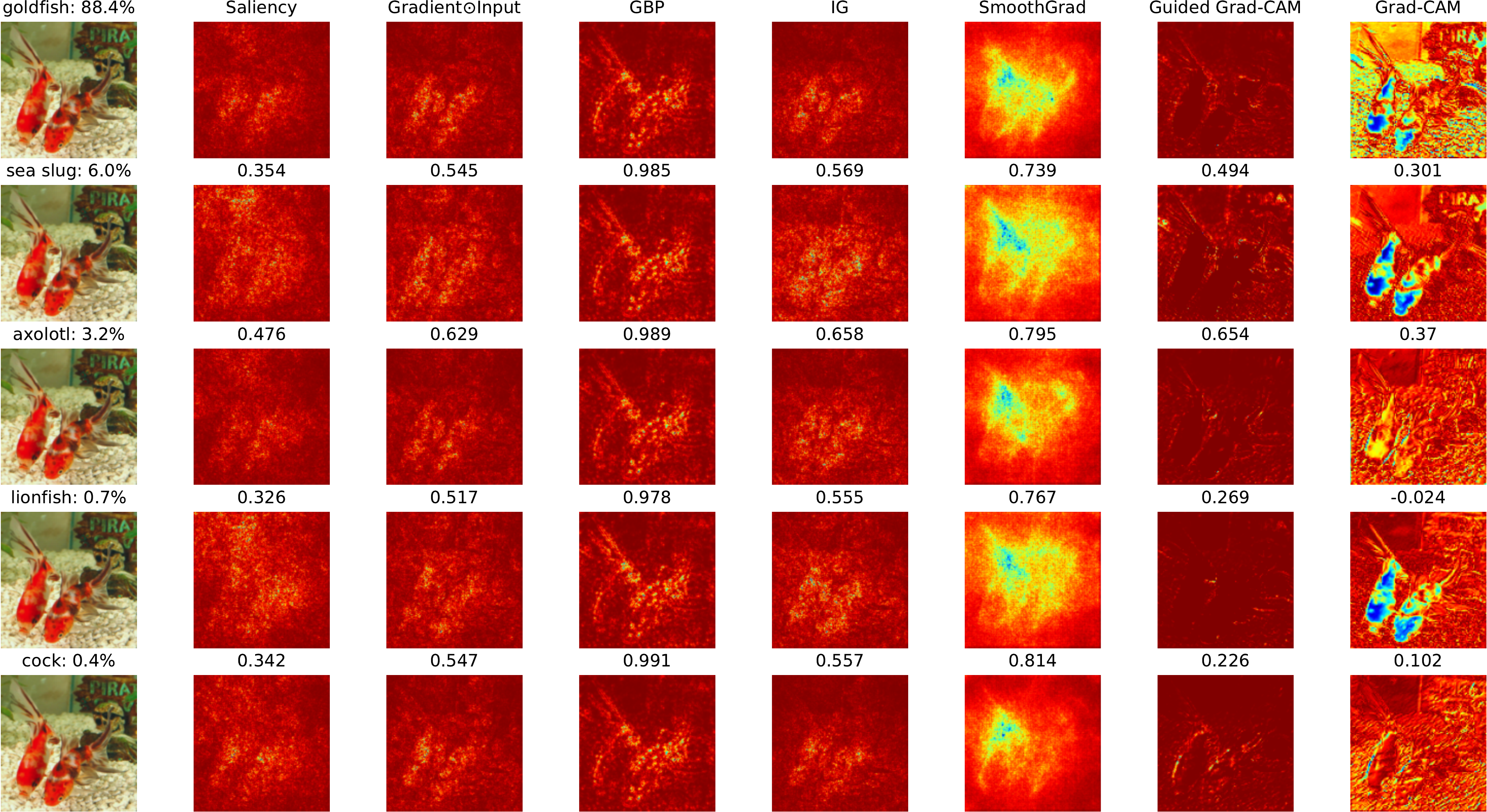}}
\caption{The class-invariant behavior of attribution methods is presented. The first column present input image and other columns show attribution maps generated by different methods. The target class and soft-max probability values are shown on the top of image in the first column. The number on the top of attribution maps are Spearman rank correlation values, calculated between the attribution maps of the true class (top row) and the target class. High correlation values show that the method is not class discriminatory, i.e., the attribution maps for any choice of class label are correlated.}
\label{fig:vis_exp}
\end{figure*}

\paragraph{Class-Invariant Behaviour}:~In some cases, the attribution maps may remain the same regardless of the class chosen by the user to compute the gradients. That is, for a given input image, similar attribution maps are generated despite the fact that the class label is changed, e.g, the network is forced to predict a certain class as in adversarial attacks \cite{rudin2019stop}. In Fig. \ref{fig:vis_exp}, we present attribution maps corresponding to 7 different methods generated for different target classes using the same input image. It is evident that most of these methods, except Grad-CAM, are not class sensitive. A similar behavior was observed when neural networks were trained using a dataset with permuted class labels. Many state-of-the-art attribution methods (except saliency maps and SmoothGrad) generated explanations that were insensitive to the permuted class labels. In summary, these methods are not able to capture the relationship between network input and output, and thus generate the same explanations, even if the class labels are changed.


\paragraph{Insensitivity to Model Parameters}:~Attribution maps should be sensitive to the learned optimal network parameters. That is, if the parameters of a trained network are replaced by random numbers, attribution maps should capture the effect of this change. However, Adebayo \emph{et al.} found that GBP and Guided Grad-CAM were insensitive to the learned parameters in the top layers (near to the output) \cite{adebayo2018sanity}. 

\paragraph{Sensitivity to Input Transformations}:~The explanations may be sensitive to factors that do not contribute to the model prediction, e.g., a constant shift in the input \cite{kindermans2019reliability}. Gradient$\odot$Input and other methods (e.g., IG) that use input in the computation of attributions are generally sensitive to such input transformations. For the case of IG, this may further depend on the chosen input baseline \cite{kindermans2019reliability}. Saliency maps, DeconvNet, and GBP were found to be insensitive to such transformations as these methods rely solely on the network parameters (no multiplication by the input) to generate attribution maps.

\paragraph{Input Dominance}:~Gradient$\odot$Input, DeepLIFT, and IG multiply the input with gradients to leverage the information present in the input features. This may help reduce noise in the attribution maps and produce more human interpretable explanations. However, in some cases, the attribution maps generated by these methods may be dominated by the input. The input does not depend on the network and cannot capture how the network processed data to make a decision \cite{adebayo2018sanity}.  

\paragraph{Partial Input Recovery}:~GBP and DeconvNet can be considered as variants of saliency maps with different rules governing negative gradients at ReLUs. These methods are able to generate relatively more human-interpretable visualizations due to the backward ReLU (used by both GBP and DeconvNet) and the local connections in CNNs. Nie \emph{et al.} showed that both GBP and DeconvNet performed (partial) input image recovery, a phenomenon that is unrelated to the network decisions \cite{nie2018theoretical}.

\paragraph{Input Baseline}:~DeepLIFT and IG use an input baseline to improve attributions. A reasonable choice of baseline depends upon the domain and task at hand. An uninformed and inappropriate choice of baseline may invalidate the explainability provided by the attribution method \cite{kindermans2019reliability}.  

\paragraph{Sensitivity to Hyperparameters}:~The explanations generated by some methods, e.g., SmoothGrad or IG may depend on the chosen hyperparameters, e.g., the number of samples used \cite{bansal2020sam}. 
\section{Explainability and Robustness}
Until recently, the explainability of DL models and their robustness were being studied in isolation \cite{tsipras2018robustness_at_odds}. However, recent work has provided a strong link between these two apparently disparate aspects of DL models. Before we explore these ideas any further, it is important to highlight two types of robustness in explainability, i.e., (1) the robustness of attribution maps, and (2) the robustness of DL models to adversarial attacks, which is intrinsically linked to their explainability.


\subsection{Attributional Robustness}
Attributional robustness is related to the stability of an attribution map in the face of a small perturbation in the input caused by natural reasons (e.g., data distribution shift) or introduced by an adversary \cite{ghorbani2019interpretation, Dombrowski_explanations, lim2021building}. It was shown that the input can be adversarially manipulated to change the attribution maps without affecting network performance, i.e., the prediction of the network does not change  \cite{ghorbani2019interpretation, Dombrowski_explanations, lim2021building}. Recent research attributes the origin of these false and manipulated explanations to the vulnerabilities of the neural network, e.g, non-smooth decision boundaries, and not the attribution generation methods \cite{Dombrowski_explanations, lim2021building}. 

\begin{figure*}[h]
\centering
\centerline{\includegraphics[height=0.28\textwidth]{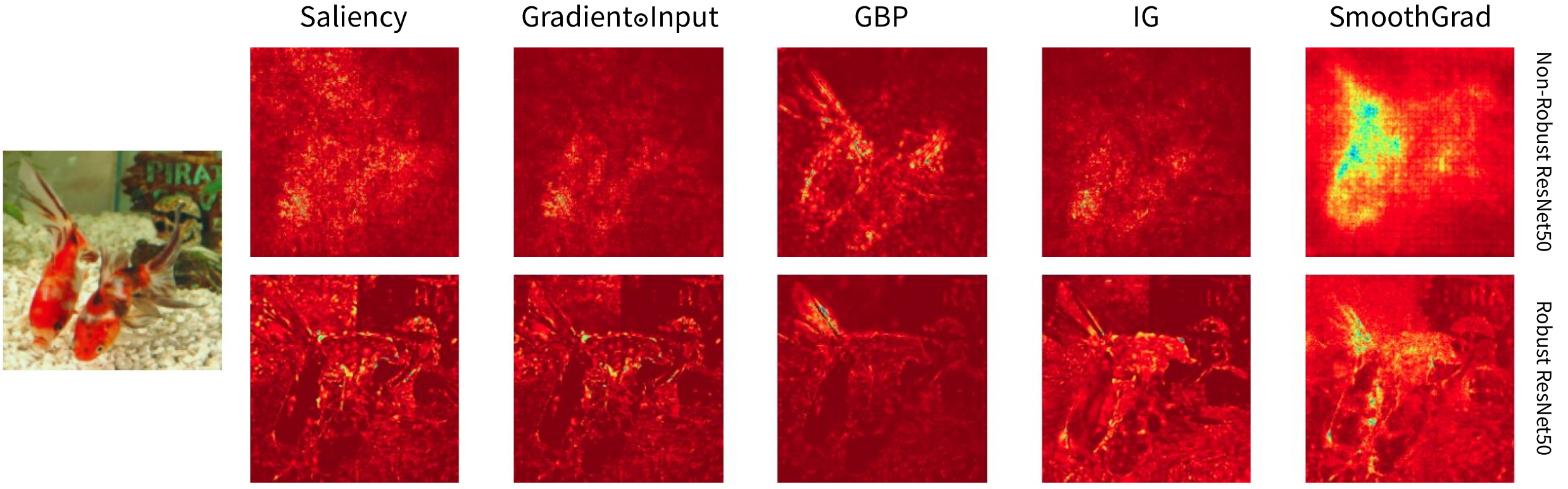}}
\caption{Input image and saliency maps generated for two different network are presented. (Left) Input image. (Top row) ResNet50 trained on natural dataset. (Bottom row) ResNet50 robustly trained using Projected Gradient Descent (PGD) attacks \cite{tsipras2018robustness_at_odds}. It is evident that attribution maps generated for adversarially trained network are more visually appealing.}
\label{fig:vg_robust_v_non}
\end{figure*}

\subsection{Adversarial Robustness}
Neural networks are known to be vulnerable to ``smart noise'' or adversarial attacks. These attacks are quasi-imperceptible perturbations in the input, measured using $L_p$ norms, that force a network to change its output \cite{madry2018towards}. %
Currently, adversarial training is the most common strategy that may provide limited defence against known attacks. In adversarial training, a modified objective function is optimized which helps in adversarial robustness by increasing the level of perturbation required to successfully change the network decision \cite{madry2018towards}. With $ \boldsymbol\theta $ denoting the learnable parameters of the model, training data $(x, y) \sim D$, and perturbation $\delta \in \Delta$, adversarial training can be formulated as the following min-max optimization problem:  
\begin{align}
    \min_{\boldsymbol\theta} \mathbb{E}_{x,y \sim D} \left[ \max_{\delta \in \Delta} \mathcal{L}(x + \delta, y; \boldsymbol\theta) \right],
\end{align}
where $\mathcal{L}$ denotes the model's loss function. 

The adversarial training can be considered as a method for the model to learn certain ($l_p$-bounded) invariances to the dataset. Some recent studies have established that learning certain types of invariances qualitatively (visually) and quantitatively may improve attribution maps, i.e., maps look more relevant to the object as viewed by a human operator \cite{tsipras2018robustness_at_odds, etmann2019connection}. In a way, adversarial training helps the network learn more like the human visual system learns. Figure \ref{fig:vg_robust_v_non} shows the difference between attribution maps generated for adversarially and naturally trained network. It is evident that the attribution maps produced by adversarially trained models seem more visually aligned with human perception \cite{ross2018improving, tsipras2018robustness_at_odds, kim2019bridging, etmann2019connection}. Tsipras \emph{et al.} described this relationship between adversarial robustness and enhanced visual alignment of attribution maps as an ``unexpected benefit'' of adversarial training \cite{tsipras2018robustness_at_odds}. Later, a number of studies verified and made an effort to explain the natural connection between adversarial robustness and explainability \cite{etmann2019connection, kim2019bridging, ignatiev2019relating}. 

Etmann \emph{et al.} showed that the improved interpretability of the saliency maps of a robustified neural network was not a side-effect of adversarial training, but a general property enjoyed by networks that are robust to adversarial perturbations \cite{etmann2019connection}. The authors showed that robustness could be defined as the distance of a test point to its closest decision boundary, and that increasing the distance (robustness) resulted in an increased alignment between the input and its attribution map. 

Recently, Kim \emph{et al.} showed that the gradients from adversarially trained networks were better aligned with the human visual system as the adversarial training caused the gradients to lie closer to the image manifold \cite{kim2019bridging}. They also reported differences in the attribution maps generated with robust networks trained using $l_2$ and $l_\infty$ adversarial images. The neural networks trained with $l_2$ were more effective at emphasizing important features while attributions from $l_\infty$-trained networks were better at identifying less important features.

Ignatiev \emph{et al.} performed a theoretical analysis using a generalized form of hitting set duality to relate explanations and adversarial examples  \cite{ignatiev2019relating}. The authors proposed the dual concept of counterexamples (and adversarial examples) and the notion of breaking an explanation. They established that each explanation must break every counterexample and vice versa. Thus, concluding that the more counterexamples (adversarial examples) the model explains, the better the interpretability of the model.

\section{Best Practices for the Community}
Explainability of black box machine learning models is important for multiple reasons, including understanding the internal workings of these models. Attribution methods are in themselves a set of mathematical operations with certain assumptions and may add another layer of abstraction over the goal of understating data and making predictions. While analysing explanations, it is also not clear how to disentangle errors in the explanation method from errors in the DL model. Currently, there is no consensus on which methods are better than others at explaining network predictions. However, there are some considerations that should be made when choosing attribution methods.

\paragraph{Gradient-based vs. Perturbation Methods}:~Gradient-based methods are computationally less expensive as in some cases these may require only one forward and one backpropagation step for estimating attributions. Perturbation-based methods generally solve an optimization problem and thus may require multiple forward passes through the network. Furthermore, gradient-based methods are more robust to input perturbations as compared to perturbation-based methods and should be preferred when robustness is a priority for the user \cite{alvarez2018robustness}.  

\paragraph{Efficiency}:~The efficiency of an attribution method can be related to the number of passes (forward and backward) through the network. Saliency maps, Input$\odot$Gradient, GBP, and Grad-CAM require one forward and one backpropagation step. IG and SmoothGrad may require $50$ to $200$ steps depending upon the problem domain, dataset, and the scope of explanation.


\paragraph{Input Baseline}:~Some attribution methods require an input baseline, which acts as the absence of the feature from the input. The baseline can be zero (a black image), an average value calculated from the dataset, a blurred version of the input image, or random values generated with Gaussian, uniform or other distributions. The choice of baseline can significantly alter the explanation \cite{sturmfels2020visualizing, kindermans2019reliability}. Since there is no current consensus on which baseline is optimal, it is difficult to recommend the use of these methods as an accurate way to explain model predictions.

\paragraph{Human Interpretability}:~Relying solely on visual analysis for understanding and comparing attribution maps can be unreliable \cite{adebayo2018sanity}. Some attribution maps may seem visually appealing, but they may not actually help us interpret model predictions. On the other hand, there is still no consensus in the community on the reliability of the various metrics to use for comparing attribution maps. Finally, a typical attribution method tends to consider each pixel as the fundamental unit explanation, which is not the basic unit used in human perception. Some recent studies have pointed out a limited usefulness of the current model explanation methods (e.g., heatmaps) and highlighted the need for a deeper investigation into the methods for presenting interpretations of models to human operators \cite{chu2020visual}. 




\section{Conclusion and Future Research Directions}
We have presented an overview of post hoc gradient-based attribution methods to explain the decisions made by DNNs. These techniques are a small but very significant part of a large body of methods that focus on explaining black box ML models. These methods are fast and can provide explanations that are robust as compared to other approaches. In some cases, these explanations may seem convincing. However, they should be approached with caution due to the inherent limitations of these methods as discussed above. Application of these methods in real-world settings, without comprehending their limitations, can create a false sense of confidence in ML decision-making. We consider that the robustness of interpretability methods is tightly coupled with the robustness of the models being explained. This area needs research efforts on both fronts, empirical as well as theoretical. There is a need to bring research communities from explainability and robustness together to explore these questions. Finally, given the state-of-the-art in post hoc explainability methods and their vulnerabilities and limitations, there is a need in the ML community to focus on building models that are inherently explainable, but as versatile, efficient, accurate and scalable as deep neural networks.  



\begin{singlespace}

\bibliographystyle{ieeetr}
\bibliography{refs.bib}

\end{singlespace}

\end{document}